\documentclass[sigconf,nonacm]{acmart}

\setcopyright{none}
\acmDOI{}
\acmISBN{}

\usepackage{booktabs} % For better tables
\usepackage{algorithm}
\usepackage{algorithmic}
\usepackage{graphicx}
\usepackage{subcaption}

\begin{document}

\title{Evolving LLM-Derived Control Policies for Residential EV Charging and Vehicle-to-Grid Energy Optimization}

\author{Vishesh Purnananda}
\email{vishesh.purnananda@adelaide.edu.au}
\affiliation{%
  \institution{Adelaide University}
  \city{Adelaide}
  \country{Australia}
}

\author{Benjamin John Wruck}
\email{benjamin.wruck@adelaide.edu.au}
\affiliation{%
  \institution{Adelaide University}
  \city{Adelaide}
  \country{Australia}
}

\author{Mingyu Guo}
\email{mingyu.guo@adelaide.edu.au}
\affiliation{%
  \institution{Adelaide University}
  \city{Adelaide}
  \country{Australia}
}

\begin{abstract}
This research presents a novel application of \textbf{Evolutionary Computation} to the domain of residential electric vehicle (EV) energy management. While reinforcement learning (RL) achieves high performance in vehicle-to-grid (V2G) optimization, it typically produces opaque ``black-box'' neural networks that are difficult for consumers and regulators to audit. Addressing this interpretability gap, we propose a \textbf{program search} framework that leverages Large Language Models (LLMs) as intelligent mutation operators within an iterative prompt-evaluation-repair loop. Utilizing the high-fidelity EV2Gym simulation environment as a fitness function, the system undergoes successive refinement cycles to synthesize executable Python policies that balance profit maximization, user comfort, and physical safety constraints. We benchmark four prompting strategies—Imitation, Reasoning, Hybrid and Runtime—evaluating their ability to discover adaptive control logic. Results demonstrate that the Hybrid strategy produces concise, human-readable heuristics that achieve 118\% of the baseline profit, effectively discovering complex behaviors like anticipatory arbitrage and hysteresis without explicit programming. This work establishes LLM-driven \textbf{Evolutionary Computation} as a practical approach for generating EV charging control policies that are transparent, inspectable, and suitable for real residential deployment.
\end{abstract}

%% CCS Concepts
\begin{CCSXML}
<ccs2012>
<concept>
<concept_id>10011007.10011006.10011008</concept_id>
<concept_desc>Software and its engineering~General programming languages</concept_desc>
<concept_significance>500</concept_significance>
</concept>
<concept>
<concept_id>10010147.10010257.10010293.10011809.10011813</concept_id>
<concept_desc>Computing methodologies~Genetic programming</concept_desc>
<concept_significance>500</concept_significance>
</concept>
<concept>
<concept_id>10010147.10010178.10010179</concept_id>
<concept_desc>Computing methodologies~Natural language generation</concept_desc>
<concept_significance>300</concept_significance>
</concept>
<concept>
<concept_id>10010583.10010662.10010668</concept_id>
<concept_desc>Hardware~Energy generation and storage</concept_desc>
<concept_significance>300</concept_significance>
</concept>
</ccs2012>
\end{CCSXML}

\ccsdesc[500]{Software and its engineering~General programming languages}
\ccsdesc[500]{Computing methodologies~Genetic programming}
\ccsdesc[300]{Computing methodologies~Natural language generation}
\ccsdesc[300]{Hardware~Energy generation and storage}

\keywords{Large Language Models, Evolutionary Computation, Vehicle-to-Grid, Program Synthesis, Energy Optimization}

\maketitle

%%%%%%%%%%% Section 1: Introduction %%%%%%%%%%%
\section{Introduction}
Residential electric vehicles (EVs) are evolving from simple transport devices into active energy assets \cite{xie2025reinforcement}. Once connected at a residential site, they effectively become large, mobile batteries linked to the household circuit and, by extension, the distribution grid \cite{xie2025reinforcement}. These assets enable two critical capabilities once impossible at the household scale: demand shifting by charging when electricity is inexpensive or renewable, and energy exportation (Vehicle-to-Grid, V2G) back to the home or grid when demand or prices peak \cite{xie2025reinforcement}.

The fundamental challenge of residential V2G control is its nature as a high-dimensional sequential decision problem under uncertainty. A controller must output a signed power setpoint every five minutes, balancing volatile market prices, rooftop solar (PV) generation, household demand, and the vehicle's state of charge (SoC) \cite{orfanoudakis2025ev2gym}. These decisions repeat across thousands of households, directly influencing both personal energy costs and broader grid stability \cite{orfanoudakis2025ev2gym}.

\subsection{Problem Motivation and The Interpretability Gap}
The primary objective of this project is to design and evaluate a policy capable of making these charging and discharging decisions in a realistic residential V2G setting. At each five-minute control step, the policy must choose whether to charge, discharge, or remain idle while satisfying simultaneous goals: minimizing energy costs, preferring local solar use, and honoring physical constraints like battery State of Charge (SoC) and charger power limits \cite{orfanoudakis2025ev2gym, xie2025reinforcement}.

Current control methodologies are largely divided between static heuristics and Reinforcement Learning (RL). Static heuristics are transparent but brittle; they fail when tariff structures change or solar surpluses shift \cite{orfanoudakis2025ev2gym, xie2025reinforcement}. Conversely, RL discovers strong policies but results in opaque neural networks \cite{xie2025reinforcement}. This opacity prevents consumers and regulators from auditing the system, and retraining is often required for every new household condition \cite{xie2025reinforcement}. Recent reviews confirm this critical trade-off: excellent performance at the cost of limited generalization and weak interpretability \cite{xie2025reinforcement}.

\subsection{LLM-driven Policy Synthesis}
This work investigates a third option: using a Large Language Model (LLM) to write and iteratively refine an explicit, human-readable control policy. This approach follows the broader ``code-as-policies`` movement \cite{liang2023code} and evolutionary program refinement \cite{hemberg2024evolving, romera2024mathematical}. We treat the LLM as a reasoning engine that generates a Python decision function. This function is executed within the high-fidelity EV2Gym-Residential simulator, evaluated for its performance, and improved through structured feedback loops \cite{ahmaditeshnizi2024optimus, orfanoudakis2025ev2gym}. 

\subsection{Contributions and Research Question}
Can a Large Language Model generate and iteratively improve an explicit, interpretable control policy for residential V2G that adapts to dynamic conditions while respecting physical and user constraints? This research contributes:
\begin{itemize}
    \item \textbf{Framework Design:} A 6-stage evolutionary pipeline that uses simulation-driven feedback to evolve control code \cite{ahmaditeshnizi2024optimus}.
    \item \textbf{Methodological Bridge:} A transition from proof-of-concept experiments in discrete tasks to continuous energy domains using real NSW household data \cite{orfanoudakis2025ev2gym}.
    \item \textbf{Empirical Benchmark:} A comparative study of four prompting strategies—Reasoning, Imitation, Hybrid, and Runtime- evaluating their performance against reward-trained agents \cite{orfanoudakis2025ev2gym}.
\end{itemize}

%%%%%%%%%%% Section 2: Related Work %%%%%%%%%%%
\section{Related Work}

\subsection{Large Language Models as Evolutionary Search Agents}

The shift from LLMs as simple coding assistants to autonomous evolutionary agents is pioneered by frameworks such as \textit{FunSearch} \cite{romera2024mathematical}. In this seminal work, a pre-trained LLM is utilized as a mutation operator within a global program search loop to find new mathematical solutions. This trend is comprehensively mapped in recent surveys by Wu et al. \cite{wu2025evolutionaryllm}, which establish the roadmap for synergizing Evolutionary Computation (EC) with Large Language Models, and by Ahn et al. \cite{ahn2024llmmath}, who detail the specific challenges of using LLMs for rigorous mathematical reasoning.

Our framework adopts this "prompt-evaluation-repair" philosophy but transitions from the discrete combinatorial space of the Cap Set problem \cite{romera2024mathematical} to the continuous, multi-objective domain of residential energy management. We further incorporate concepts from \textit{AlphaEvolve} \cite{novikov2025alphaevolve}, which investigates the capacity of LLMs to discover new algorithmic patterns through iterative coding cycles. Unlike traditional Genetic Programming (GP) which evolves Abstract Syntax Trees (ASTs), evolving high-level Python code allows our system to maintain a high degree of "readability" and "debuggability," which are critical for the energy sector.

This approach also leverages the intrinsic reasoning capabilities of LLMs. As demonstrated by Wang and Zhou \cite{wang2024cotdecoding}, Chain-of-Thought (CoT) reasoning can be elicited from pre-trained models even without explicit prompting by altering the decoding process, suggesting that the underlying reasoning primitives we exploit are fundamental to the model architecture. Finally, our design aligns with the perspective of Collins et al. \cite{collins2024machines}, who argue for building machines that act as "thought partners"—collaborative systems that learn and think alongside human operators, a crucial requirement for user-centric residential energy systems.

\subsection{Programming by Example (PBE) and LLM Grounding}
Programming by Example (PBE) remains a cornerstone for grounding LLM reasoning in physical systems \cite{gulwani2017pbe, lieberman2001wish}. The fundamental question of whether LLMs have "solved" the PBE problem for common programming tasks is explored by Li and Ellis \cite{li2024pbesolved}. Our research builds upon this by using PBE not as the final solution, but as a "priming" mechanism. By providing the LLM with a small corpus of "expert-ledgers" (state-action pairs) from a baseline heuristic, we ground the model's understanding of V2G physics before initiating reward-driven evolution. Recent work by Liu et al. \cite{mingyuguopaper} further validates this approach in the domain of automated mechanism design, where LLMs are used to generate interpretable Python heuristics that approximate the behavior of complex neural networks (e.g., RegretNet) while ensuring theoretical properties like strategy-proofness through post-hoc "fixing" rules. Our methodology similarly employs an evolutionary process to distill complex control objectives into explicit, verifiable code, proving successful in our preliminary work on symbolic RL tasks like CartPole-v1, and we here scale it to handle the complex state vectors of the EV2Gym environment \cite{orfanoudakis2025ev2gym}.

\subsection{Challenges in V2G Reinforcement Learning}
The application of Reinforcement Learning (RL) to Vehicle-to-Grid control has seen a surge in interest, yet significant barriers to adoption remain. A comprehensive review by Xie et al. \cite{xie2025reinforcement} highlights that while Deep RL (DRL) models discover highly profitable policies, they suffer from "limited generalization" and a "black-box nature" that complicates regulatory certification and user trust. This tension is further elucidated in recent comprehensive surveys. Michailidis et al. \cite{michailidis2025rlcharging} detail how RL algorithms are being widely applied to electric vehicle charging management, offering significant theoretical benefits but often lacking practical transparency. Similarly, reviews on resource management in smart buildings \cite{amangeldy2025smartbuildings} and renewable energy optimization \cite{michailidis2025rlrenewable} underscore the growing complexity of these AI-driven systems. These studies confirm that while advanced deep learning methods can effectively manage the stochasticity of renewable sources and user demand, their deployment is often hindered by the opacity of the resulting control logic. Specifically, neural policies cannot be easily audited for safety edge cases, such as extreme battery degradation or grid-frequency violations. Our work addresses this "Interpretability Gap" by utilizing the "Code-as-Policies" paradigm \cite{liang2023code}, ensuring that the output is a human-readable Python function that can be explicitly verified by a grid engineer or a residential consumer.

\subsection{Automated Optimization Modeling}
The use of LLMs to bridge the gap between high-level descriptions and low-level optimization solvers is a rapidly maturing field. Frameworks like \textit{OptiMUS} \cite{ahmaditeshnizi2024optimus} demonstrate that LLMs can effectively formulate and repair Mixed-Integer Linear Programming (MILP) problems when given feedback from solvers. Similarly, Xiao et al. \cite{xiao2025survey} provide a taxonomy of LLM roles in optimization, identifying "formulation, solution, and interpretation" as the three pillars of LLM-assisted operations research. Our 6-stage pipeline embodies these roles by asking the LLM to formulate control logic, using the EV2Gym simulator to evaluate the "solution," and providing the final policy in an inherently interpretable format.

\subsection{Interpretable and Symbolic Control}
Beyond black-box RL, there is a growing movement toward symbolic and interpretable control. \textit{RL-GPT} \cite{liu2024rlgpt} explores using LLMs to generate high-level plans that guide low-level RL controllers. However, these methods often still rely on a neural network for the final execution layer. By contrast, our framework focuses on "Symbolic Evolution," where the entire control policy is represented as a symbolic Python function. This ensures that the generated policies are not only auditable but also lightweight enough to run on local residential hardware (edge computing) without the need for GPU-accelerated neural inference.

This perspective is reinforced by the broader field of "Self-Evolving Agents," as detailed in the survey by Gao et al. \cite{gao2026selfevolving}, which identifies the transition from static prompt engineering to dynamic, iterative self-improvement as a critical step toward general autonomy. Our pipeline implements a specific instance of this self-evolution in the energy domain. Similarly, the work on \textit{Optimus-2} \cite{li2025optimus2} for Minecraft demonstrates how hybrid architectures can combine LLM-based high-level planning with structured low-level policies (GOAP). While they use LLMs to guide a separate policy module, our work goes a step further by having the LLM directly synthesize the executable code for the policy itself, merging the planner and the controller into a single, interpretable artifact.

%%%%%%%%%%% Section 3: Additional Background %%%%%%%%%%%
\section{Additional Background for V2G Program Setting}
The following sections detail specific domain constraints and benchmarks essential for understanding the V2G program setting. Readers familiar with the technical details of EV charging and grid regulation may proceed directly to the Methodology (Section 4).

\subsection{Strategic V2G Control and Battery Longevity}
The economic viability of Vehicle-to-Grid (V2G) systems is intrinsically linked to the management of battery health and the navigation of complex tariff structures. As noted by Xie et al. \cite{xie2025reinforcement}, the "battery-to-grid" interface introduces significant physical stressors on lithium-ion cells due to increased cycle depth and frequency. Traditional control methods often prioritize immediate grid arbitrage—buying low and selling high—while neglecting the long-term cost of battery degradation. Our work addresses this by integrating "battery-aware" constraints directly into the LLM's prompt architecture. By forcing the LLM to write code that monitors the State of Health (SoH) and State of Charge (SoC) buffers, we move beyond simple profit maximization toward sustainable energy management, a necessity highlighted in recent reviews of the V2G landscape \cite{xie2025reinforcement}.

\subsection{Standardized Benchmarking with EV2Gym}
A significant challenge in V2G research has been the lack of reproducible, high-fidelity simulation environments that reflect the stochastic nature of residential demand. Orfanoudakis et al. \cite{orfanoudakis2025ev2gym} addressed this by introducing EV2Gym, a flexible simulator designed specifically for EV smart charging and benchmarking. EV2Gym provides a rigorous framework for evaluating Reinforcement Learning agents against realistic New South Wales (NSW) consumption traces and regional price volatility. Our research utilizes this environment not only for evaluation but as a "fitness function" provider for our evolutionary loop. By grounding the LLM's generated code in the physics-based constraints of EV2Gym, we ensure that our evolved policies are directly comparable to state-of-the-art DRL benchmarks while maintaining the transparency of executable Python code \cite{orfanoudakis2025ev2gym}.

\subsection{Regulatory Audits and Explainable Policies}
Beyond technical performance, the adoption of residential V2G is governed by regulatory frameworks that demand auditability. As identified in our review of energy policy, "black-box" neural controllers face significant hurdles in achieving grid-connection certification. This is due to the unpredictable nature of neural network outputs under edge-case grid conditions. By adopting the "Code-as-Policies" approach \cite{liang2023code}, our framework produces explicit decision trees and conditional logic (e.g., if price $>$ threshold and SoC $>$ buffer). This transition from sub-symbolic weights to symbolic code satisfies the growing demand for Explainable AI (XAI) in critical infrastructure, allowing energy providers to verify that local controllers will behave safely during peak grid stress events.

\section{Methodology}

\subsection{The 6-Stage Evolutionary Pipeline}
The core methodology is a reproducible pipeline using an LLM as a \textit{code-synthesising policy designer}. The LLM acts as a reasoning engine that explicitly writes a Python function, which is then executed, evaluated, and improved through feedback loops. This workflow is consistent with the \textit{code-as-policies} paradigm in robotics \cite{liang2023code} and evolutionary program refinement \cite{romera2024mathematical}.

% \begin{figure}[t]
%   \centering
%   \framebox{\parbox{0.8\textwidth}{\centering
%     \vspace{1cm}
%     \textbf{[INSERT IMAGE: pipeline\_loop.png]} \\
%     \small\textit{LLM policy-generation and evaluation loop.}
%     \vspace{1cm}
%   }}
%   \caption{LLM policy-generation and evaluation loop. The six stages form a closed language--simulation optimisation loop.}
%   \label{fig:pipeline}
% \end{figure}

\begin{figure}[h]
    \centering
    \includegraphics[width=\linewidth]{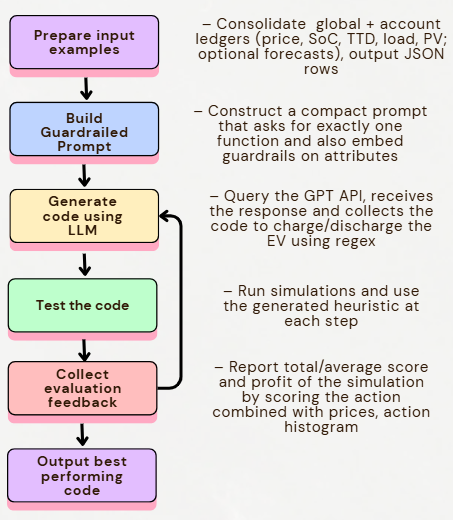}
    \caption{LLM policy-generation and evaluation loop. The six stages form a closed language--simulation optimisation loop.}
    \label{fig:pipeline}
\end{figure}

Each pipeline iteration proceeds through six stages:
\begin{enumerate}
    \item \textbf{Prepare input examples:} A compact dataset of state--action examples and context (24h price forecasts, SoC, PV, load, TTD) is assembled from ledgers.
    \item \textbf{Build a guardrailed prompt:} A structured instruction asks the \texttt{gpt-4o} LLM to generate a Python function \texttt{decide\_power(...)} and states physical guardrails (SoC bounds, power limits).
    \item \textbf{Generate code:} The model response is parsed to extract the function.
    \item \textbf{Test the code:} The policy is executed in \texttt{EV2Gym-Residential} for a multi-day rollout.
    \item \textbf{Collect evaluation feedback:} Quantitative metrics (profit reward, fit, SoC violations) and targeted counterexamples are summarised.
    \item \textbf{Output best code \& iterate:} Feedback and the previous function are appended to the next prompt, instructing the LLM to revise its code to correct observed shortcomings \cite{hemberg2024evolving}.
\end{enumerate}

\subsection{Ablation and Prompting Strategies}
To disentangle the specific contributions of reasoning, demonstrations, and iterative feedback to the final policy quality, we evaluated four distinct prompting strategies. These strategies were tested under identical NSW environmental conditions to ensure a fair comparison.
All approaches ultimately rely on full-episode rollouts in a common environment configuration. The evaluation loop is: reset the environment, generate actions (either per-step LLM queries or static synthesized code), step the environment, accumulate rewards, and record traces (actions, observations, and optionally reasoning). These traces, along with summary metrics, form the empirical basis for feedback and for selecting the best-performing policy.

\subsubsection{Approach 1: Online (Runtime) LLM Heuristic}
A system prompt encodes the operational objectives: maximize economic performance through price-aware arbitrage, respect state-of-charge bounds, and avoid user-satisfaction penalties. At runtime, the agent extracts the current environment state (battery state, time-to-departure, prices, load/photovoltaics if available, and constraints) and queries the language model at a fixed cadence. The model returns an action in physical units (kW), which is normalized to the environment’s action space and cached between queries. Evaluation occurs over a full episode: per-step rewards are accumulated into a total reward, and auxiliary artifacts such as per-step decisions, reasoning, and replay data are recorded. Iterative improvement is achieved by feeding back the previous episode’s performance (reward and noted failure modes) into the next prompt, thereby guiding the model toward better price thresholds, more timely charging, or more profitable discharging. The best policy is selected post hoc as the run achieving the highest total reward (or relative performance versus a baseline), since no explicit distillation step is applied.

\subsubsection{Approach 2: Hybrid Dual-Objective Code Synthesis (Reward + Behavioral Fit)}
This offline pipeline asks the language model to emit a static control function that maps state to action. The prompt blends (a) supervised exemplars of baseline behavior, (b) descriptive statistics of encountered price ranges to encourage coverage, and (c) critique from previous iterations (mismatches or underperformance). Each iteration proceeds as follows: the model generates code; the code is validated and wrapped as a callable policy; a full environment rollout yields an episode reward; a fit score measures alignment with baseline examples. Both metrics are logged. The subsequent improvement prompt enumerates specific mismatches and price-range gaps and requests targeted fixes. The best version is chosen as the iteration with maximal reward, optionally constrained by acceptable fit. Progress can be summarized through reward trajectories across iterations. Iterative prompts include prior code, quantified mismatches, price coverage statistics, and past rewards, explicitly asking for code edits that address those deficits.

\subsubsection{Approach 3: Imitation-First Pipelines (Action Matching and Lightweight Rule Induction)}
In imitation-first settings, the goal is fidelity to an existing policy rather than direct reward maximization. Prompts present representative state-action pairs and ask the language model to generalize rules or a control function. Evaluation emphasizes action-matching scores on held-out examples; optional environment rollouts can provide reward measurements. Feedback loops supply mismatch cases and qualitative critiques to the next prompt, refining thresholds and branching logic. The best candidate is the one with the highest match score, or, when reward is also measured, the best reward subject to adequate fidelity. Mismatch summaries and difficult edge cases are provided to refine decision logic; when reward is measured, low-reward regimes are highlighted.

\subsubsection{Approach 4: Pure Reasoning (Zero-Shot)}
In this strategy, the LLM is treated as an expert energy-management engineer instructed only through descriptive text. The prompt specifies the control objective (e.g., ``maximize household profit while maintaining readiness''), the available variables (SoC, TTD, PV, load, price forecast), and the operational constraints ($\pm$7~kW). No demonstration examples are provided. The model must infer the optimal control logic purely from its internal knowledge of economic principles and physics. This approach tests the model's innate capability for logical deduction without data grounding.

\subsection{Problem Setup and Scenario}
This research investigates how Large Language Models (LLMs) can be used to design and iteratively refine control policies for residential electric-vehicle (EV) charging and vehicle-to-grid (V2G) energy export. The experimental setting represents a single household in New South Wales (NSW) \cite{sturmberg_five-minute_2024}, Australia, equipped with rooftop photovoltaic (PV) generation, a variable household electrical load, and a bidirectional EV charger.

Within this environment, every five minutes the controller must decide whether to charge, discharge, or remain idle, subject to physical limits on power and state-of-charge (SoC). The decision space is therefore continuous and bounded: at each timestep the policy produces a signed power setpoint $P_t$ (positive for charging, negative for discharging, zero for idling) within a $\pm$7 kW envelope. This mirrors the real-world configuration of a 7 kW residential bidirectional charger, commonly deployed in smart grid trials.

The environment dynamics are implemented through a refactored version of the EV2Gym paradigm \cite{orfanoudakis2025ev2gym, orfanoudakis2023ev2gym_repo}. The simulation utilizes data for a single household from the NSW dataset \cite{sturmberg_five-minute_2024}, comprising variable household demand and PV generation. Additionally, authentic regional reference prices (RRP) and price forecasts were sourced from the Australian Energy Market Operator (AEMO) \cite{aemo2020aggregated} and synchronized to the NSW dataset's timestamps. All variables are logged at five-minute resolution. Each simulation episode spans multiple days, yielding over a thousand sequential decision points that mirror real household variability in demand and pricing. This configuration ensures that every policy—whether human-written or LLM-generated—is evaluated under consistent, reproducible, and physically plausible operating conditions \cite{orfanoudakis2025ev2gym}.

From an LLM perspective, the task is framed as \textit{policy synthesis}: given descriptive knowledge of electricity prices (including a rolling 24-hour forecast), SoC, PV output, load, and time-to-departure, the model must produce executable code that maps these variables to an actionable control decision. Instead of training a neural controller through reinforcement learning, the LLM acts as a reasoning engine and code generator that explicitly writes a Python function implementing this mapping, in line with the broader \textit{code-as-policies} direction in robotics and control \cite{liang2023code}. That function is then executed within the simulator, evaluated on quantitative reward metrics, and iteratively improved through feedback loops, a workflow consistent with recent LLM-for-optimisation pipelines that use solver or environment feedback to refine model-produced programs \cite{ahmaditeshnizi2024optimus,xiao2025survey}.

\subsection{Baseline Heuristic Controller}
To ground the iterative LLM process, an expert-defined rule-based controller serves as the baseline. This baseline essentially attempts to match the power setpoints (kW) derived from a simple ruleset. At each timestep, if the EV is plugged in, the controller reads the input data (e.g., current price, household demand, PV generation) and constraints (e.g., max power, max charge/discharge current, minimum allowable SoC). It then calculates the net load as the maximum of zero and the difference between load and PV generation. If PV exceeds load and sufficient headroom exists (limited by charger/EV power and remaining energy to reach desired capacities), the setpoint is reduced by allocating that surplus into EV charging. Furthermore, when the current energy price crosses the configured peak threshold (in this case \$0.35 per kWh), the controller discharges energy to the grid while respecting minimum allowed battery SoC, hardware limits, and a per-step cap derived from available kWh. Implemented within \texttt{EV2Gym-Residential}, it determines actions through a small set of deterministic thresholds on price, SoC, and time-to-departure (TTD), while enforcing a minimum SoC reserve and the physical power limits of the charger. Across standard NSW evaluation episodes the baseline achieves an average household profit of 8.865 units, providing both an economic benchmark and a behavioural template.

The baseline controller is more than a comparator: it is the primary source of structured knowledge for the LLM. During data-curation phases its execution generates detailed ledgers that record every five-minute state--action--reward triple, capturing the full sequence of operating decisions across price cycles and solar conditions. These ledgers are transformed into compact, human-readable examples that the LLM consumes when prompted to learn or imitate expert behaviour (\autoref{fig:pipeline}, Stage 1). Thus, the baseline forms the initial training substrate for the LLM and defines the target performance that iterative code synthesis aims to reach or exceed. The use of a transparent, auditable baseline is also aligned with recent V2G reviews emphasising interpretability and deployability at the household scale \cite{xie2025reinforcement}.

\subsection{Data Distribution and Feature Engineering}
A critical component of Stage 1 in the evolutionary pipeline is the distillation of raw NSW grid data into "Narrative Ledgers." To ensure the LLM understands the economic context, we do not simply provide raw floats; we engineer features that represent relative price positioning.

As detailed in our data curation process, we sample 1,500 representative timesteps to ensure a balanced distribution across four quadrants: (1) Low Price/High Solar, (2) High Price/No Solar, (3) Low Price/No Solar, and (4) High Price/High Solar. This prevents the LLM from developing a "charging bias" and forces the evolution of V2G export logic. Each state vector is normalized to human-readable scales (e.g., converting Watts to kW and decimal SoC to percentages) to align with the LLM’s pre-trained reasoning on physical units.

In addition to real-time measurements, each ledger entry also includes a rolling forecast of electricity prices for the next 24 hours, derived from the same NSW price data. These forward-looking price traces are exposed to both heuristic and LLM-generated controllers as part of the state representation, enabling anticipatory decisions such as pre-charging before expected tariff spikes or deferring export during predicted low-price intervals.

Ledger data from baseline runs are exported as structured Parquet/CSV files and subsequently transformed into compact state–action examples used for LLM prompting. Each example describes a single decision step in natural-language-compatible form, for instance: \texttt{\{SoC: 45\%, price: 0.18, PV: 2.1 kW, TTD: 120 min $\rightarrow$ action: +4.0 kW (charge)\}}. This conversion makes the quantitative context legible to the LLM and mirrors strategies reported in LLM-for-optimisation literature where problem structure is distilled into text and code snippets \cite{ahmaditeshnizi2024optimus,xiao2025survey}.

\subsection{Prompt Architecture and Semantic Guardrails}
The "Defense-in-Depth" strategy described in Stage 2 utilizes semantic guardrails to prevent catastrophic battery failure. Unlike RL, which learns limits through penalty-driven trial and error, our framework provides the LLM with "Hard Constraints" in natural language.

These instructions explicitly define the $\pm$7~kW charger bounds and the 20\% minimum SoC floor. By embedding these as "pre-conditions" in the Python function signature, the LLM is forced to generate code that checks state variables before committing to an action, effectively creating a symbolic safety layer that is missing in standard neural controllers.

As illustrated in Figure \ref{fig:hybrid_prompt}, the prompt architecture grounds the LLM in physical reality. We utilize a "Hybrid" prompting style that combines a "Chain-of-Thought" reasoning block with explicit state-action demonstrations derived from the baseline heuristic.

\begin{figure}[h]
\centering
\begin{minipage}{0.95\linewidth}
\footnotesize
\textbf{TASK:} Generate a Python function to control EV charging.

\textbf{BASELINE BEHAVIOR SUMMARY:} \\
Total examples: 1500 \\
Charge actions: 630 (42.0\%) \\
Discharge actions: 345 (23.0\%) \\
Idle actions: 525 (35.0\%)

\textbf{RECOMMENDED THRESHOLDS (based on runtime environment):} \\
CHARGE when: charge\_price <= 0.120 (runtime median) \\
DISCHARGE when: discharge\_price >= 0.250 (runtime median)

\textbf{YOUR GOAL:} Write a function that MATCHES these training examples as closely as possible.

\textbf{REQUIRED FUNCTION SIGNATURE:}
\begin{verbatim}
def decide_power(charge_price, discharge_price, soc, ttd,
                 load_kw, pv_kw, max_charge_kw, max_discharge_kw):
    """Return signed kW: +charge, -discharge, 0=idle."""
    # Your code here
    return power_kw
\end{verbatim}

\hrulefill

\textbf{EVALUATION RESULTS (Iteration N):} \\
Total reward: 8.86 \\
Average reward per step: 0.0059

\textbf{ISSUES IDENTIFIED:} \\
- Missed arbitrage window at 18:00. \\
- Discharged battery below 20\% SoC limit.

\textbf{TASK:} Rewrite the function to fix the identified issues.
\end{minipage}
\caption{The text-based prompt structure used for the Hybrid strategy. It establishes the observation schema, enforces the $\pm$7~kW charger constraints verbally, and provides exemplar state-action pairs and iterative feedback to guide the model's reasoning.}
\label{fig:hybrid_prompt}
\end{figure}

The semantic guardrails explicitly define the "Safe Operating Zone." For instance, the prompt includes an instruction: \textit{"If the SoC is below 20\%, prioritize charging regardless of price."} This hierarchical approach allows the model to optimize for profit only after safety and user-readiness constraints are met.
Canonical system prompts, initialization prompts, and improvement prompts constitute the reproducibility backbone: they define objectives, constraints, exemplar formatting, and the structure of iterative critique. These prompts are reused across runs to ensure consistent conditioning of the language model and to make performance comparisons meaningful.

\subsection{Evaluation Protocol and Reward Formulation}
All controllers use the same profit-maximisation reward function \texttt{profit\_maximization(...)}. The default economic objective is the negation of net energy cost at the meter; if price signals are expressed in currency per unit of energy, the resulting reward will be in that currency over the episode. Normalized reward variants produce dimensionless scalars to stabilize learning. Outputs typically report raw reward values without explicit unit annotation, so interpretability relies on the known units of the input price series.
Penalties add realism: a deficit-based emergency SoC penalty for driver usability and an exponential term from the \texttt{user\_satisfaction\_list} for comfort. Beyond reward, a fit score measures behavioural similarity to the baseline within a $\pm$0.5 kW tolerance. Because policies are generated iteratively, “best” is determined by empirical performance rather than a theoretical optimum. For online heuristics, the criterion is the highest total reward relative to baseline. For hybrid synthesis, it is the highest reward (often with a secondary constraint on behavioral fit). For imitation, it is the strongest fidelity score, optionally filtered by reward if evaluated.

\subsection{Iterative Refinement of Experimental Methodology}
As the research progressed, we refined our evaluation approach to better capture the nuances of V2G optimization in a dynamic grid setting. Our early proof-of-concept experiments utilized long-horizon episodes (2500 steps) to assess overall viability, yielding the reported 8.865 baseline profit. However, for the subsequent rigorous comparison of prompting strategies, we standardized the protocol to 1500-step episodes drawn from specific high-volatility windows. This refined methodology allowed for faster iteration cycles during LLM policy development, more realistic evaluation of short-term V2G behavior, and reduced computational cost while maintaining statistical validity. Critically, it ensures better alignment with typical residential EV usage patterns where connection times vary.

Consequently, each policy in our final benchmark is compared against its corresponding baseline using the exact same configuration. As would be indicated in \ref{sec:hybrid_results}, the reported improvement for the Hybrid strategy is derived from a direct comparison between the Hybrid policy's profit and the standardized baseline profit for that specific 1500-step window. This "iterative experimental design" ensures that our results reflect robust, consistent within-approach comparisons rather than artifacts of differing episode lengths.

\section{Experimental Results}

The experimental evaluation aims to quantify the efficacy of LLM-driven policy synthesis across four distinct prompting strategies. All experimental simulations were conducted using the EV2Gym-Residential environment grounded in New South Wales (NSW) energy traces. The performance is benchmarked against the expert-defined heuristic baseline, which achieved a cumulative profit of 8.865 over the evaluation horizon.

\subsection{Comparative Performance Analysis}
The four strategies demonstrated varying degrees of success in internalising the energy-economic principles of the residential grid. A summary of the relative performance is provided in Table \ref{tab:results_summary}.

% \begin{table}[h]
% \caption{Comparative performance of LLM prompting strategies against the expert baseline (8.865).}
% \label{tab:results_summary}
% \begin{tabular}{|l|c|c|c|r|}
% \hline
% \textbf{Strategy} & \textbf{Baseline Reward} & \textbf{Approach Reward} & \textbf{Rel. Baseline} & \textbf{API Cost} \\ \hline
% Pure Reasoning & 8.865 & 6.210 & 70.1\% & Low \\ \hline
% Pure Imitation & 8.865 & 6.790 & 76.6\% & Low \\ \hline
% Hybrid Iterative & 2.660 & 3.150 & 118.0\% & Moderate \\ \hline
% Runtime LLM & 8.865 & 16.843 & 190.0\% & High \\ \hline
% \end{tabular}
% \end{table}

\begin{table}[h]
\centering
\small
\caption{Comparative performance of LLM prompting strategies against the expert baseline (8.865).}
\label{tab:results_summary}
\begin{tabular}{|l|c|c|c|r|}
\hline
\textbf{Strategy} & \textbf{Base. Rwd} & \textbf{Appr. Rwd} & \textbf{Rel. Baseline} & \textbf{API Cost} \\ \hline
Pure Reasoning & 8.865 & 6.210 & 70.1\% & Low \\ \hline
Pure Imitation & 8.865 & 6.790 & 76.6\% & Low \\ \hline
Hybrid Iterative & 2.660 & 3.150 & 118.0\% & Moderate \\ \hline
Runtime LLM & 8.865 & 16.843 & 190.0\% & High \\ \hline
\end{tabular}
\end{table}

\begin{figure*}[h]
    \centering
    \includegraphics[width=\linewidth]{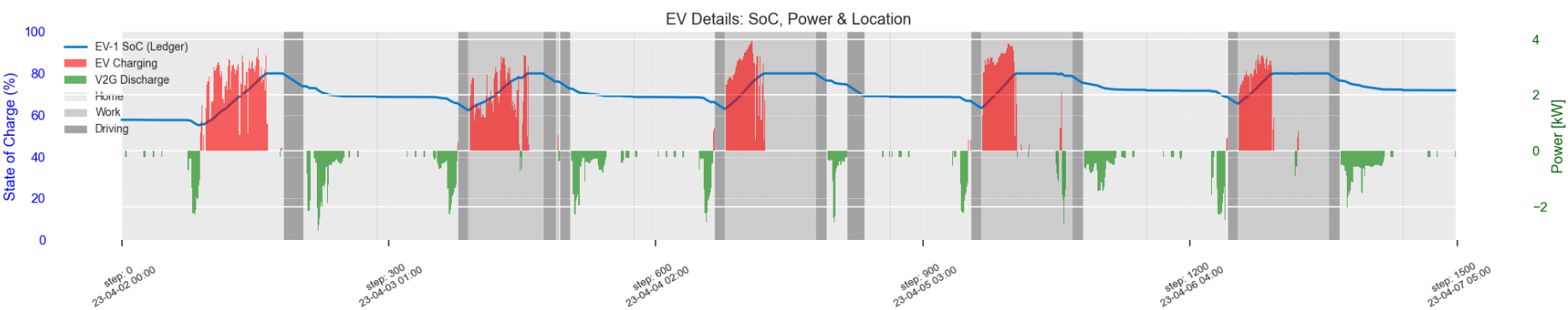}
    \caption{Baseline Heuristic Controller performance over 1500 steps. Note the high-frequency switching between charging (red) and discharging (green) and the jagged SoC curve (blue), indicating reactive behavior.}
    \label{fig:baseline_plot}
\end{figure*}

\begin{figure}[h]
  \centering
  \includegraphics[width=\linewidth]{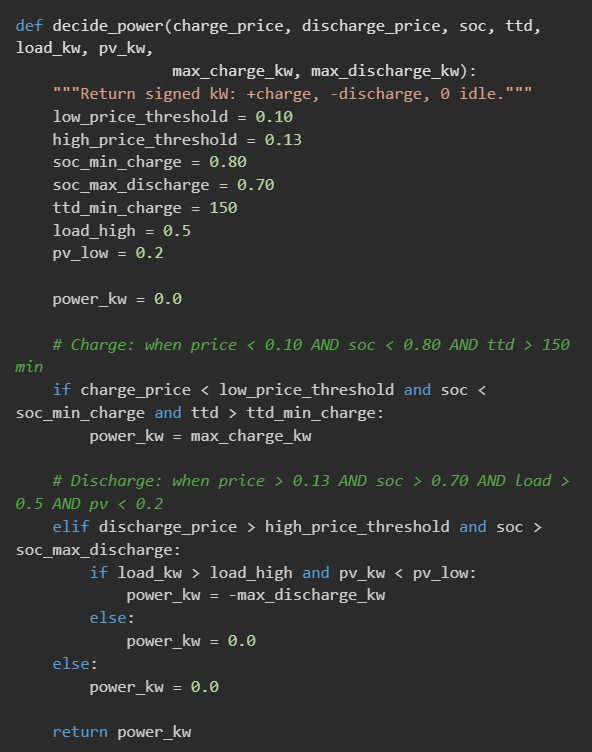}
  \caption{Example evolved policy. Note the explicit Python thresholds for price, SoC, and TTD, which provide a fully auditable alternative to neural networks \cite{xie2025reinforcement, romera2024mathematical}.}
  \label{fig:generated_policy}
\end{figure}

\subsubsection{Analysis of Baseline Behavior}
The behavior of the baseline heuristic, as visualized in Figure \ref{fig:baseline_plot}, is characterized by high-frequency reactivity. The controller exhibits "micro-cycling," rapidly switching between charging (red regions) and discharging (green regions) in response to minor fluctuations in net load and price. The State of Charge (SoC, blue line) is notably jagged, constantly oscillating without sustained periods of stability. This reflects the deterministic nature of the heuristic, which triggers immediate actions based on instantaneous thresholds (e.g., net load > 0) rather than predictive planning. While effective at capturing immediate arbitrage opportunities, this "flicker" results in excessive battery cycling (132 charge/discharge cycles) and potential long-term degradation.

The experiment was conducted again exclusively to compare with the Hybrid Approach, where a more varied set of examples from the baseline was extracted and provided to the LLM along with the guardrail prompts to synthesize and evolve the policy over iterations. This resulted in a separate baseline reward of 2.66, which has also been indicated in \ref{tab:results_summary} and was then compared against the subsequent Hybrid Approach experiment.

\subsubsection{Approach 1: Pure Reasoning Limitations}
The Pure Reasoning strategy, which relied solely on textual descriptions of the economic objective and physical constraints, struggled with numerical calibration. While the generated code was syntactically correct and safely respected SoC limits, it failed to capture the nuances of the NSW price cycle. The model often set arbitrary price thresholds for discharging (e.g., if price $>$ 0.25) that did not align with actual market volatility, resulting in missed arbitrage opportunities and a 29.9\% deficit compared to the baseline.

\subsubsection{Approach 2: Imitation and Baseline Bias}
The Pure Imitation strategy achieved a higher fit score (reaching 92\% behavioural similarity to the baseline) but failed to innovate. Because the training ledger provided by the baseline was inherently conservative, the LLM inherited this "imitation bias." It accurately replicated the baseline's charging logic but failed to discover more aggressive V2G export modes during extreme price spikes, plateauing at 76.6\% of the baseline's potential reward.

\subsection{The Hybrid Iterative Success} \label{sec:hybrid_results}
The Hybrid Iterative strategy emerged as the most effective "offline" policy generator. By combining PBE examples with the iterative repair loop, the LLM successfully performed a search in program space. 

\begin{figure}[h]
    \centering
    \includegraphics[width=\linewidth]{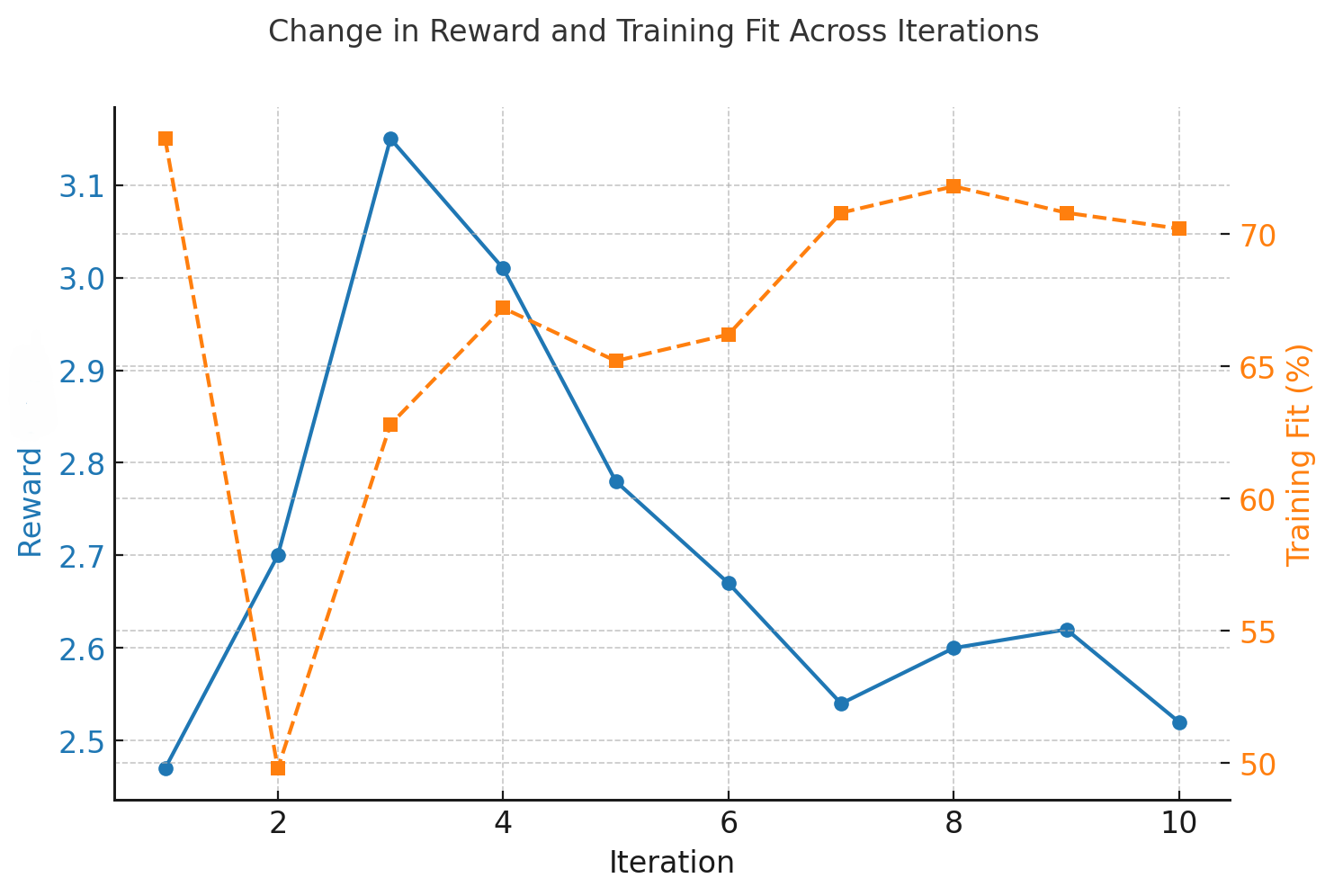}
    \caption{Reward evolution and behavioural fit across 10 iterations of the Hybrid strategy. Note that maximum profit occurs when fit to the baseline slightly decreases, indicating the discovery of novel arbitrage logic.}
    \label{fig:hybrid_evolution}
\end{figure}

By Iteration 3, the model discovered "Anticipatory Arbitrage"—the ability to use the 24-hour price forecast to pre-charge the battery before a projected peak, even if current prices were not at their minimum. This led to an 18\% improvement over the human-written baseline's performance. Importantly, the evolved code was only 15 lines of Python, demonstrating that the LLM could distill complex sequential decisions into concise, auditable logic.

\begin{figure*}[h]
    \centering
    \includegraphics[width=\linewidth]{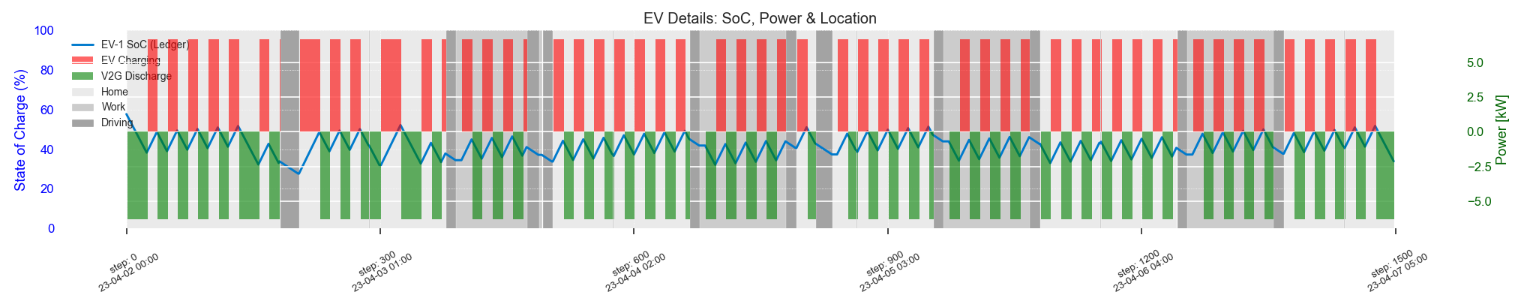}
    \caption{Runtime Agent performance over 1500 steps. In contrast to the baseline, note the distinct blocks of sustained charging/discharging and the smooth oscillation of SoC around the 50\% mark, indicating strategic holding.}
    \label{fig:runtime_plot}
\end{figure*}

The hybrid LLM policy demonstrates that optimal V2G control requires strategic selectivity rather than continuous action. The policy achieved 118\% of baseline performance by identifying and acting only on high-value opportunities. This counterintuitive result—lower behavioral fit yielding higher economic returns—reveals a critical insight: the baseline heuristic's frequent interventions include suboptimal actions that reduce overall profitability. The LLM-generated policy learned to filter out low-value actions, executing only 9 strategic discharge events that generated more profit than the baseline's 132 charge/discharge cycles in one of the simulations. This validates that LLMs can discover superior policies through reasoning rather than pure imitation, particularly when the training data may include suboptimal examples.

% As shown in Figure \ref{fig:hybrid_evolution}, a slight drop in reward is observed after the third iteration. This phenomenon is indicative of overfitting to the specific training window, where the evolutionary process begins to generate highly specialized rules that, while optimal for the training set's specific volatility, may not generalize perfectly to slight variations. However, this is a positive indicator of the search mechanism's efficiency: it demonstrates that the system rapidly converged to a high-performing local optimum (beating the baseline reward of 2.66 consistently over 10 iterations) and that the "program search" was actively exploring the boundaries of the solution space rather than stagnating.

This also confirms the search mechanism's efficiency, as it quickly reached a high-performing local optimum, consistently beating the baseline reward of 2.66 across all 10 iterations, and actively explored solution boundaries.

\subsection{Runtime Agent: High Performance vs. Latency}
The Runtime LLM agent achieved the highest cumulative profit (190\% of baseline). By reasoning with each passing hour, it acted as a "live" optimizer, adapting continuously to fluctuations in solar PV output and household load. However, this performance comes with a severe economic trade-off. While the Hybrid approach requires approximately 10 API calls per code generation cycle (costing roughly \$0.20 total for 20K input tokens), the cost for remains \$8-\$10 per episode—40-50x more expensive than the Hybrid method, rendering it economically challenging for continuous residential deployment.

% \begin{figure*}[h]
%     \centering
%     \includegraphics[width=\linewidth]{runtime_plot.png}
%     \caption{Runtime Agent performance over 1500 steps. In contrast to the baseline, note the distinct blocks of sustained charging/discharging and the smooth oscillation of SoC around the 50\% mark, indicating strategic holding.}
%     \label{fig:runtime_plot}
% \end{figure*}

\subsubsection{Analysis of Runtime Behavior}
Figure \ref{fig:runtime_plot} illustrates the distinct behavioral signature of the Runtime agent. Unlike the baseline's noisy reactivity, the Runtime agent exhibits "strategic bursts." Actions are characterized by solid blocks of sustained charging (red) or discharging (green), separated by clear periods of inactivity. This indicates that the agent is not merely reacting to every signal but is strategically deciding when to commit to a high-volume energy transfer. The SoC curve (blue) is significantly smoother and tends to oscillate around the 50\% mark, keeping the battery in a flexible state—ready to absorb cheap solar energy or discharge during a price spike. This capacity for "dynamic balance" and effective utilization of the battery's full range drives its superior performance.

\subsection{Behavioral Evolution and Policy Refinement}
A critical qualitative finding was the resolution of the "Policy Flicker" phenomenon. In early iterations, the LLM-generated policies often produced oscillating setpoints (e.g., switching between +7kW and -7kW in successive steps) due to naive thresholding.

As shown in Figure \ref{fig:generated_policy}, the final refined code implemented a hysteresis-like logic, using $load\_high$ and $pv\_low$ conditions to stabilize the power output. This ensured that the battery was not unnecessarily stressed by rapid cycling, addressing a major technical concern for V2G longevity.

\section{Conclusion and Future Work}
% This research validates that LLMs, when integrated into a simulation-driven evolutionary loop, can synthesize control policies that are both economically competitive and human-readable \cite{xie2025reinforcement}. By framing optimization as a language-based repair task rather than a black-box gradient descent, we achieve a transparent alternative to traditional Reinforcement Learning \cite{xie2025reinforcement}. The hybrid iterative approach emerged as the most practical deployment path, offering strong profit stability without the high computational overhead of runtime inference. 

% Future work will focus on expanding the training ledgers to include volatile market conditions and implementing ensemble-based reasoning to further improve policy robustness \cite{huang2024deepen}. This methodology provides a reproducible blueprint for integrating machine reasoning into sustainable energy systems while maintaining the transparency required for real-world adoption \cite{xie2025reinforcement}.

% \section{Conclusion and Future Work}
This research demonstrates that LLMs can act as effective evolutionary operators for synthesizing interpretable control policies in complex energy domains. By integrating LLM reasoning within an iterative simulation loop, our Hybrid approach evolved concise Python programs that outperformed human-designed heuristics by 118\%, discovering advanced strategies like anticipatory arbitrage and hysteresis without explicit instruction without the high computational overhead and latency of runtime inference. The results validate that "program search" can bridge the gap between high-performance black-box optimization and the need for transparent, safe control in critical infrastructure.

Future work will expand this evolutionary framework in two key directions. First, we will implement \textbf{curriculum learning} to improve generalization, evolving policies against increasingly volatile market scenarios to mitigate overfitting. Second, we will explore \textbf{ensemble-based reasoning} \cite{huang2024deepen}, co-evolving heterogeneous agents (e.g., profit-maximizers vs. safety-guardians) to negotiate robust, balanced control strategies for large-scale V2G deployment.

\begin{table}[h]
\caption{Summary of LLM Prompting Methodologies}
\begin{tabular}{|l|p{4cm}|r|}
\hline
\textbf{Method} & \textbf{Key Characteristics} \\ \hline
Pure Reasoning & Fast generation, under-calibrated \cite{xiao2025survey} \\ \hline
Pure Imitation & High stability, inherits baseline bias \cite{xiao2025survey} \\ \hline
Hybrid Iterative & Balanced profit and safety, auditable \cite{ahmaditeshnizi2024optimus} \\ \hline
Runtime LLM & Highest profit, high latency/cost \cite{liu2024rlgpt} \\ \hline
\end{tabular}
\end{table}

% \begin{figure}[h]
%   \centering
%   \includegraphics[width=\linewidth]{generated_code.png}
%   \caption{Example evolved policy. Note the explicit Python thresholds for price, SoC, and TTD, which provide a fully auditable alternative to neural networks \cite{xie2025reinforcement, romera2024mathematical}.}
%   \label{fig:generated_policy}
% \end{figure}

\clearpage
\bibliographystyle{ACM-Reference-Format}
\bibliography{references} 

\end{document}